\documentclass{article}
\usepackage{spconf,amsmath,graphicx}
\usepackage{multicol}
\usepackage{multirow}
\usepackage{booktabs}
\usepackage{rotating}
\usepackage[small]{caption}
\usepackage[table]{xcolor}
\usepackage{hyperref}

\def\ours{TR3D}
\def\oursff{TR3D+FF}

\newcommand\inline{\noindent\textbf}
\newcommand\bt{\textbf}

\title{TR3D: Towards Real-Time Indoor 3D Object Detection}
\name{Danila Rukhovich \qquad Anna Vorontsova \qquad Anton Konushin}
\address{
Samsung Research\\
}

\makeatletter
\def\blfootnote{\xdef\@thefnmark{}\@footnotetext}
\makeatother

\begin{document}
%
\maketitle

\begin{abstract}

Recently, sparse 3D convolutions have changed 3D object detection. Performing on par with the voting-based approaches, 3D CNNs are memory-efficient and scale to large scenes better. However, there is still room for improvement. With a conscious, practice-oriented approach to problem-solving, we analyze the performance of such methods and localize the weaknesses. Applying modifications that resolve the found issues one by one, we end up with \ours{}: a fast fully-convolutional 3D object detection model trained end-to-end, that achieves state-of-the-art results on the standard benchmarks, ScanNet v2, SUN RGB-D, and S3DIS. Moreover, to take advantage of both point cloud and RGB inputs, we introduce an early fusion of 2D and 3D features. We employ our fusion module to make conventional 3D object detection methods multimodal and demonstrate an impressive boost in performance. Our model with early feature fusion, which we refer to as \oursff, outperforms existing 3D object detection approaches on the SUN RGB-D dataset. Overall, besides being accurate, both \ours{} and \oursff{} models are lightweight, memory-efficient, and fast, thereby marking another milestone on the way toward real-time 3D object detection. Code is available at \url{https://github.com/SamsungLabs/tr3d}.

\end{abstract}
\begin{keywords}
3D object detection, indoor scene understanding, point clouds
\end{keywords}
\section{Introduction}
\label{sec:intro}


The recent emergence of self-driving, AR/VR applications, 3D modeling, and household robotics attracted attention to 3D object detection as a core scene understanding technology. 

Modern 3D object detection methods can be categorized into voting-based, transformer-based, and 3D convolutional. Voting-based methods process points with a feature extractor network, use center votes to create an object proposal, and accumulate point features within each group. Many voting-based methods have poor scalability, limiting their usage. 

Instead of domain-specific heuristics and hyperparameters, transformer-based methods use end-to-end learning and forward pass on inference. Being more generalized, they still have issues when processing larger scenes. 

3D convolutional methods represent point clouds as voxels, which allows processing sparse 3D data efficiently. Dense volumetric features require a lot of memory, so sparse representation and sparse 3D convolutions are used instead. Compared to other methods, 3D sparse convolutional methods are memory-efficient and scale to large scenes well without sacrificing point density. Up until very recently, such methods lacked accuracy, yet due to the recent advances in the field, fast and scalable yet accurate methods were developed~\cite{rukhovich2022fcaf3d}.

Overall, modern 3D detection methods demonstrate impressive results with only geometric inputs; yet, possibilities of leveraging data of other modalities for 3D object detection have been investigated as well. While point clouds are difficult to obtain and require additional equipment, RGB cameras are much more accessible. Typically being incorporated into capturing devices, they provide cheap, easy-to-use yet extremely informative data. 
Existing approaches add RGB data in late stages; either they have a complicated, memory-consuming architecture~\cite{rukhovich2021imvoxelnet}, rely on custom procedures limiting their usage~\cite{park2021multimodal}, or run slow iterative schemes~\cite{wang2022multimodal}. On the contrary, we present an \textit{early fusion strategy}, which can also be integrated into other point cloud-based models. Combined with the proposed simple yet efficient fully-convolutional pipeline, this strategy allows achieving state-of-the-art results in 3D object detection from point cloud and RGB data.

\section{Related Work}
\label{sec:related}

\inline{3D object detection in point clouds.} Voting-based methods pioneered the field, with VoteNet~\cite{qi2019votenet} being the first method that introduced point voting for 3D object detection. VoteNet extracts features from 3D points, assigns a group of points to each object candidate according to their voted center, and computes object features from each point group. BRNet~\cite{cheng2021brnet} refines voting results with the representative points from the vote centers, which improves capturing the fine local structural features. 
H3DNet~\cite{zhang2020h3dnet} improves the point group generation procedure by predicting a hybrid set of geometric primitives. 
RBGNet~\cite{wang2022rbgnet} proposes a ray-based feature grouping module, which aggregates the point-wise features on object surfaces by uniformly emitting rays from cluster centers. 

\inline{Transformer-based methods.} In transformer-based methods, grouping is not guided with a set of hyperparameters explicitly, which makes them less domain-specific. GroupFree \cite{liu2021group-free} employs a transformer module to update object query locations iteratively and accumulate intermediate results. 3DETR~\cite{misra20213detr} was the first to solve the 3D object detection task with a transformer model.

\inline{Voxel-based methods.} Voxel-based 3D object detection methods~\cite{rukhovich2022fcaf3d, hou20193dsis, gwak2020gsdn} convert points into voxels and process them with 3D convolutional networks. However, dense volumetric features still consume much memory. GSDN~\cite{gwak2020gsdn} alleviates this issue using sparse 3D convolutions, yet being notably less accurate than modern voting-based methods. Unlike GSDN~\cite{gwak2020gsdn}, FCAF3D~\cite{rukhovich2022fcaf3d} does not utilize anchors, and addresses 3D object detection in a data-driven manner, making it the first voxel-based approach that performs on par with state-of-the-art voting-based methods. In our work, we use FCAF3D as a strong baseline.

\inline{Point cloud and RGB fusion.} 
Semantic data contained in images might provide additional clues for 3D object detection. 
Recent works~\cite{qi2020imvotenet, huang2020epnet, liu2021epnetcb} generate initial region proposals and then refine them, using RGB features as guidance. ImVoteNet~\cite{qi2020imvotenet} leverages 2D detection results to perform voting. 
EPNet++~\cite{liu2021epnetcb} fuses image features with intermediate outputs of a feature extraction model. 
MMTC~\cite{park2021multimodal} uses cascades, including a 2D segmentation network between the first and second stages of a 3D object detection network. TokenFusion~\cite{wang2022multimodal} fuses point cloud and RGB information with a transformer model and learns to replace uninformative tokens with projected and aggregated inter-modal features. 

\section{Proposed Method}
\label{sec:method}

\ours{} is based on FCAF3D~\cite{rukhovich2022fcaf3d} and inherits its simple, fully-convolutional design (Fig.~\ref{fig:scheme}), yet with some updates improving its efficacy and accuracy. Moreover, we propose an early feature fusion for 3D object detection from RGB and point clouds, and incorporate a fusion module into \ours{} to construct a multimodal \oursff{} model.

\subsection{\ours: 3D Object Detection Method}
\label{ssec:3d-object-detection}

\begin{figure}[t!]
    \begin{center}
        \includegraphics[width=\linewidth]{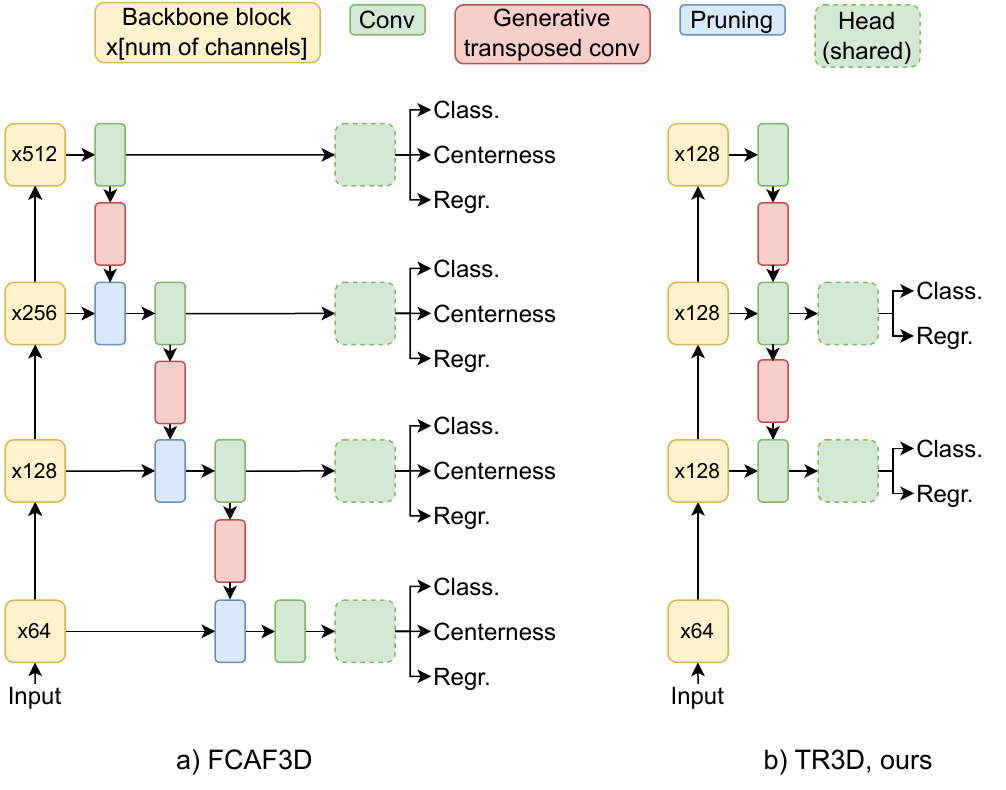}
    \end{center}
    \caption{Scheme of TR3D in comparison with the FCAF3D baseline.}
\label{fig:scheme}
\end{figure}

Considering FCAF3D~\cite{rukhovich2022fcaf3d} as a baseline, we introduce several modifications. \ours{} is a result of all these modifications being applied jointly (Fig.~\ref{fig:scheme}). To ensure these modifications contribute to the final performance in the desired way, we apply them gradually, one at a time, and estimate the detection accuracy on S3DIS, memory footprint, and inference speed (hereinafter denoted as FPS, it is actually the number of scenes processed per second). We report the results of this series of experiments in Tab.~\ref{tab:ablation-components}.

\inline{Efficacy.} First, we aim to turn the baseline model into a fast and lightweight one. The examination of performance revealed that a single generative transposed convolutional layer in the head at the first level consumes as much as one-third of the total memory. Accordingly, we drop the head on the first level: not only it has a major impact on the memory footprint, which decreases $1.5$ times from $661$ to $415$ Mb, but also adds $+6$ FPS. Without this head, the pruning layer appears to be redundant, so we omit it. Then, we remove the head at the fourth level, since it focuses on processing large objects, which are rare in indoor scenes. Furthermore, if restricting the number of output channels in the backbone blocks, the number of parameters gets reduced dramatically from $68.3$ to $14.7$ M, and the memory consumption halves. 

\begin{figure*}[h!]
    \begin{center}
        \includegraphics[width=0.975\linewidth]{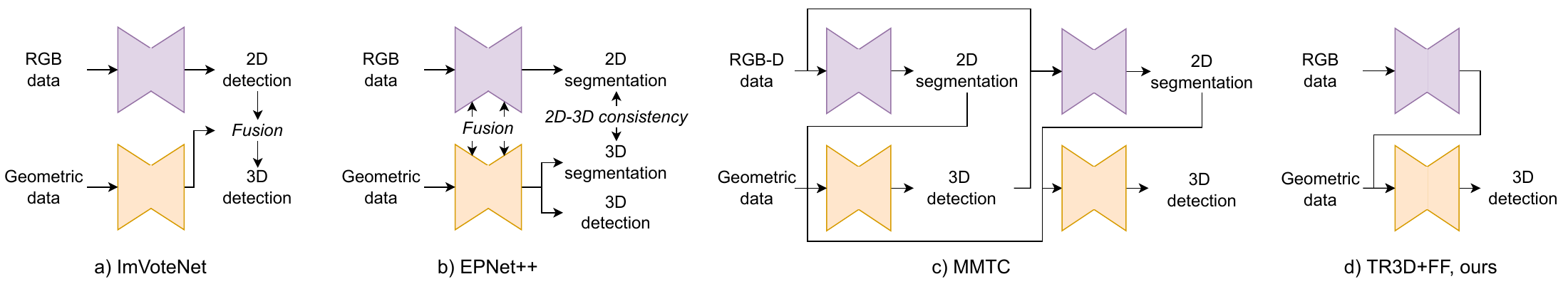}
    \end{center}
    \caption{Our early feature fusion strategy and existing RGB and point cloud fusion approaches: ImVoteNet~\cite{qi2020imvotenet}, EPNet++\cite{liu2021epnetcb}, MMTC\cite{park2021multimodal}. With its simple and straightforward design, it appears to be more beneficial in terms of detection accuracy.
    }
\label{fig:feature-fusion}
\end{figure*}

\begin{table}[h!]
\centering 
    \setlength{\tabcolsep}{3pt}
    \resizebox{\linewidth}{!}{
    \begin{tabular}{lcccc}
    \toprule
    \multirow{2}{*}{Modification} & \multirow{2}{*}{mAP} & \multirow{2}{*}{FPS} & \#Params, & Memory,\\
    & & & M & Mb \\ 
    \midrule
    Baseline~\cite{rukhovich2022fcaf3d} & 66.7 (64.9) & \cellcolor{blue!10}10.9 & 70.5 & \cellcolor{blue!10}661\\
    \hspace*{2mm}- head at 1st level & 62.9 (61.3) & \cellcolor{blue!10}16.9 & 70.1 & \cellcolor{blue!10}415\\
    \hspace*{2mm}- pruning & 62.9 (61.3) & 17.3 & 70.1 & 415 \\
    \hspace*{2mm}- head at 4th level & 61.5 (59.6) & 17.3 & \cellcolor{blue!10}68.3 & 408\\
    \hspace*{2mm}- \textgreater 128 channels & 61.2 (58.8) & 20.8 & \cellcolor{blue!10}14.7 & 207\\
    \hspace*{2mm}- centerness & \cellcolor{blue!10}61.5 (59.8) & 21.0 & 14.7 & 207\\
    \hspace*{2mm}+ TR3D assigner & \cellcolor{blue!10}72.9 (71.4) & 21.0 & 14.7 & 207\\
    \hspace*{2mm}+ DIoU loss & \bt{74.5 (72.1)} & \bt{21.0} & \bt{14.7} & \bt{207}\\
    \bottomrule
    \end{tabular}
    }
    \caption{Results of a study of the TR3D components on S3DIS. Blue cells mark the most significant gains. Our approach has a 3x smaller memory footprint, has 4.5x fewer parameters, and is almost 2x faster on inference.}
    \label{tab:ablation-components}
\end{table}

\inline{Accuracy.} In the second round of improvements, we mainly focus on detection accuracy. Our experiments demonstrate that centerness does not contribute to the prediction quality, and can be harmlessly skipped. FCAF3D assigner considers only points inside 3D bounding boxes. Accordingly, this imposes a risk of missing thin or small objects (e.g., whiteboards), which might fall between locations and hence not get assigned with ground truth boxes. So, we introduce a \ours{} assigner considering not inside points but the nearest ones, which might be located outside bounding boxes as well. Moreover, we pre-define the head level for each object category: typically large objects (e.g., \textit{bed} or \textit{sofa}) are processed at the third level, and smaller ones (e.g., \textit{chair} or \textit{night stand}) are handled at the second. Switching to a new multi-level assigner boosts the performance from $61.5$ to $72.9$ mAP. With a novel assigner, the assigned points might be outside the ground truth bounding box. Accordingly, IoU might be equal to zero, so, to enforce the training process, we replace IoU loss with DIoU loss, which resolves such cases successfully. This final update allows achieving $74.5$ mAP at $21$ FPS, with a lightweight model with $14.7$M parameters and a peak memory consumption of $207$ Mb. Overall, \ours{} consumes 3x less memory than the baseline, has 4.5x fewer parameters, and requires 2x less time to proceed.

\subsection{\oursff: RGB and Point Cloud Fusion}
\label{ssec:feature-fusion}

\ours{} performs early 2D-3D feature fusion. First, RGB images are processed with a frozen ResNet50+FPN network pretrained to solve a 2D object detection task. Then, extracted 2D features are projected into 3D space, same as in~\cite{rukhovich2021imvoxelnet}. Finally, the fusion is performed by summing up projected 2D features with 3D features element-wise. This strategy is notably more simple, time- and memory-efficient, than existing approaches (Fig.~\ref{fig:feature-fusion}); surprisingly, it ensures better results.

We incorporate our fusion module into VoteNet~\cite{qi2019votenet} and \ours, and evaluate the performance in point cloud-based and multimodal setting. In addition, we compare VoteNet+FF with ImVoteNet~\cite{qi2020imvotenet} implementing another feature fusion strategy on top of VoteNet~\cite{qi2019votenet}. According to the metrics provided in Tab.~\ref{tab:fusion}, our early feature fusion boosts the detection accuracy of VoteNet by +6.8 mAP@0.25, while ImVoteNet demonstrates a smaller gain of 5.7 mAP@0.25. At the same time, the detection accuracy of \ours{} is getting improved by +2.3 mAP@0.25 and +3.0 mAP@0.5 with feature fusion. This slight yet steady improvement evidences geometric data being the major source of information, while visual data is complementary, serving as guidance to alter the estimates. 


\begin{table}[h!]
\centering 
    \resizebox{\linewidth}{!}{
    \begin{tabular}{lcccc}
    \toprule
    Method & Inputs & mAP@0.25 & mAP@0.5 \\ \midrule
    VoteNet\cite{qi2019votenet} & PC & 57.7 & - \\
    ImVoteNet\cite{qi2020imvotenet} & PC+RGB & 63.4 & - \\
    VoteNet+FF, ours & PC+RGB & 64.5 (63.7) & 39.2 (38.1) \\
    \midrule
    \ours, ours & PC & 67.1 (66.3) & 50.4 (49.6) \\
    \oursff, ours & PC+RGB & \bt{69.4} (68.7) & \bt{53.4} (52.4) \\
    \bottomrule
    \end{tabular}
    }
    \caption{Results of point cloud-based 3D object detection methods against multimodal methods with our early fusion, on SUN RGB-D.}
    \label{tab:fusion}
\end{table}

\section{Experiments}
\label{sec:experiments}

\subsection{Experimental Settings}
\label{ssec:setup}

\inline{Datasets.} The experiments are conducted on SUN RGB-D~\cite{song2015sunrgbd}, ScanNet v2~\cite{dai2017scannet}, and S3DIS~\cite{armeni2016s3dis}.
Richly-annotated ScanNet v2~\cite{dai2017scannet} contains 1513 reconstructed scans: 1201 comprise the training subset and 312 are left for validation. We calculate axis-aligned bounding boxes of 18 object categories from semantic per-point annotation, as proposed in ~\cite{qi2020imvotenet}. SUN RGB-D~\cite{song2015sunrgbd} is a monocular dataset with 10 355 RGB-D images. The training and validation subsets contain 5285 and 5050 point clouds, respectively, with annotated oriented bounding boxes. We consider the 10 most common object categories for evaluation, as proposed in \cite{qi2019votenet}. S3DIS ~\cite{armeni2016s3dis} features 271 scenes within 6 large areas. Following the standard evaluation protocol, we assess detection accuracy on scans from Area 5, using 5 semantic categories. 

\inline{Metrics.} For all datasets, we use mean average precision (mAP) under IoU thresholds of 0.25 and 0.5 as a metric. To eliminate outliers caused by randomness and to obtain statistically significant results, we train our models five times and evaluate each trained model five times independently. For a fair comparison with existing methods that follow the same protocol, we report both the best and the average value (in brackets) across 25 trials for each metric. 

\begin{table*}[ht!]
\centering \setlength{\tabcolsep}{2.5pt}
    \resizebox{0.975\linewidth}{!}{
    \begin{tabular}{llccccccccc}
    \toprule
    \multirow[l]{2}{*}{Method} & \multirow[l]{2}{*}{Presented at} & \multicolumn{3}{c}{ScanNet} & \multicolumn{3}{c}{SUN RGB-D} & \multicolumn{3}{c}{S3DIS} \\
    \cmidrule(lr){3-5} \cmidrule(lr){6-8} \cmidrule(lr){9-11}
    & & mAP@0.25 & mAP@0.5 & FPS & mAP@0.25 & mAP@0.5 & FPS & mAP@0.25 & mAP@0.5 & FPS \\ \midrule
    VoteNet\cite{qi2019votenet} & ICCV'19 & 58.6 & 33.5 & 14.1 & 57.7 & - & 24.9 & - & - & - \\
    GSDN\cite{gwak2020gsdn} & ECCV'20 & 62.8 & 34.8 & - & - & - & - & 47.8 & 25.1 & - \\
    BRNet\cite{cheng2021brnet} & CVPR'21 & 66.1 & 50.9 & \textless14 & 61.1 & 43.7 & \textless24 & - & - & - \\
    3DETR\cite{misra20213detr} & ICCV'21 & 65.0 & 47.0 & \textless7 & 59.1 & 32.7 & \textless12 & - & - & - \\
    H3DNet\cite{zhang2020h3dnet} & ECCV'20 & 67.2 & 48.1 & 7.9 & 60.1 & 39.0 & \textless24 & - & - & - \\
    GroupFree \cite{liu2021group-free} & ICCV'21 & 69.1 (68.6) & 52.8 (51.8) & 8.3 & 63.0 (62.6) & 45.2 (44.4) & \textless24 & - & - & - \\
    RBGNet~\cite{wang2022rbgnet} & CVPR'22 & 70.6 (69.9) & 55.2 (54.7) & \textless14 & 64.1 (63.6) & 47.2 (46.3) & \textless24 & - & - & - \\
    HyperDet3D~\cite{zheng2022hyperdet3d} & CVPR'22 & 70.9 & 57.2 & \textless14 & 63.5 & 47.3 & \textless24 & - & - & - \\
    FCAF3D~\cite{rukhovich2022fcaf3d} & ECCV'22 & 71.5 (70.7) & 57.3 (56.0) & 15.7 & 64.2 (63.8) & 48.9 (48.2) & 17.9 & 66.7 (64.9) & 45.9 (43.8) & 10.9 \\
    \bt{\ours, ours} & - & \bt{72.9} (72.0) & \bt{59.3} (57.4) & \bt{23.7} & \bt{67.1} (66.3) & \bt{50.4} (49.6) & \bt{27.5} & \bt{74.5} (72.1) & \bt{51.7} (47.6) & \bt{21.0} \\
    \bottomrule
    \end{tabular}
    }
    \caption{Results of \ours{} and existing 3D object detection methods based on geometric inputs only. \ours{} outperforms the strong baseline FCAF3D in all benchmarks.}
    \label{tab:results}
\end{table*}

\inline{Implementation details.} 
Our models are implemented using a mmdetection3d framework~\cite{mmdet3d2020} and trained and tested on a single NVidia 4090 GPU.  
We follow the training procedure of FCAF3D~\cite{rukhovich2022fcaf3d}, using the same losses, optimizer, learning schedule, and augmentations. 

\subsection{Comparison to Prior Work}
\label{ssec:comparison}

\inline{Point cloud-based 3D object detection.} We report quantitative results of our \ours{} on ScanNet v2, SUN RGB-D, and S3DIS in Tab.~\ref{tab:results}. As can be observed, \ours{} demonstrates a solid superiority in all benchmarks, in terms of all metrics. The most notable accuracy gain is achieved for S3DIS, where the difference is as large as +7.8 mAP@0.25 and +5.8 mAP@0.5. Overall, we claim our approach to set a new state-of-the-art in indoor 3D object detection based on only geometric inputs. 

\inline{Multimodal 3D object detection.} The results of multimodal methods on SUN RGB-D, including \oursff, are listed in Tab.~\ref{tab:results}. \oursff{} surpasses previous state-of-the-art MMTC~\cite{park2021multimodal} by 4.1 mAP@0.25 and 4.8 mAP@0.5.

\begin{table}[h!]
\centering \setlength{\tabcolsep}{3pt}
    \resizebox{\linewidth}{!}{
    \begin{tabular}{llccc}
    \toprule
    Method & Presented at & mAP@0.25 & mAP@0.5 & FPS \\
    \midrule
    ImVoteNet~\cite{qi2020imvotenet} & CVPR'20 & 63.4 & - & 14.8 \\
    EPNet~\cite{huang2020epnet} & ECCV'20 & 64.6 & - & - \\
    TokenFusion~\cite{wang2022multimodal} & CVPR'22 & 64.9 (64.4) & 48.3 (47.7) & -\\
    EPNet++~\cite{liu2021epnetcb} & TPAMI'22 & 65.3 & - & -\\
    MMTC~\cite{park2021multimodal} & BMVC'21 & 65.3 (64.7) & 48.6 (48.2) & \textless7\\ 
    \bt{\oursff, ours} & - & \bt{69.4} (68.7) & \bt{53.4} (52.4) & \bt{17.5}\\
    \bottomrule
    \end{tabular}
    }
    \caption{Results of \oursff{} and existing 3D object detection methods using both point clouds and RGB, on SUN RGB-D. \oursff{} is a state-of-the-art in multimodal 3D object detection.}
    \label{tab:results-ff}
\end{table}

\subsection{Qualitative Results}
\label{ssec:qualitative}

The input point clouds, ground truth and predicted bounding boxes are depicted in Fig.~\ref{fig:qualitative}.

\begin{table}[h!]
    \centering
    \setlength{\tabcolsep}{1pt}
      \begin{tabular}{rcc}
        & Ground truth & Predicted \\
        \parbox[t]{2mm}{\rotatebox[origin=l]{90}{\hspace*{7mm}Scannet}} & 
        \includegraphics[width=0.475\linewidth]{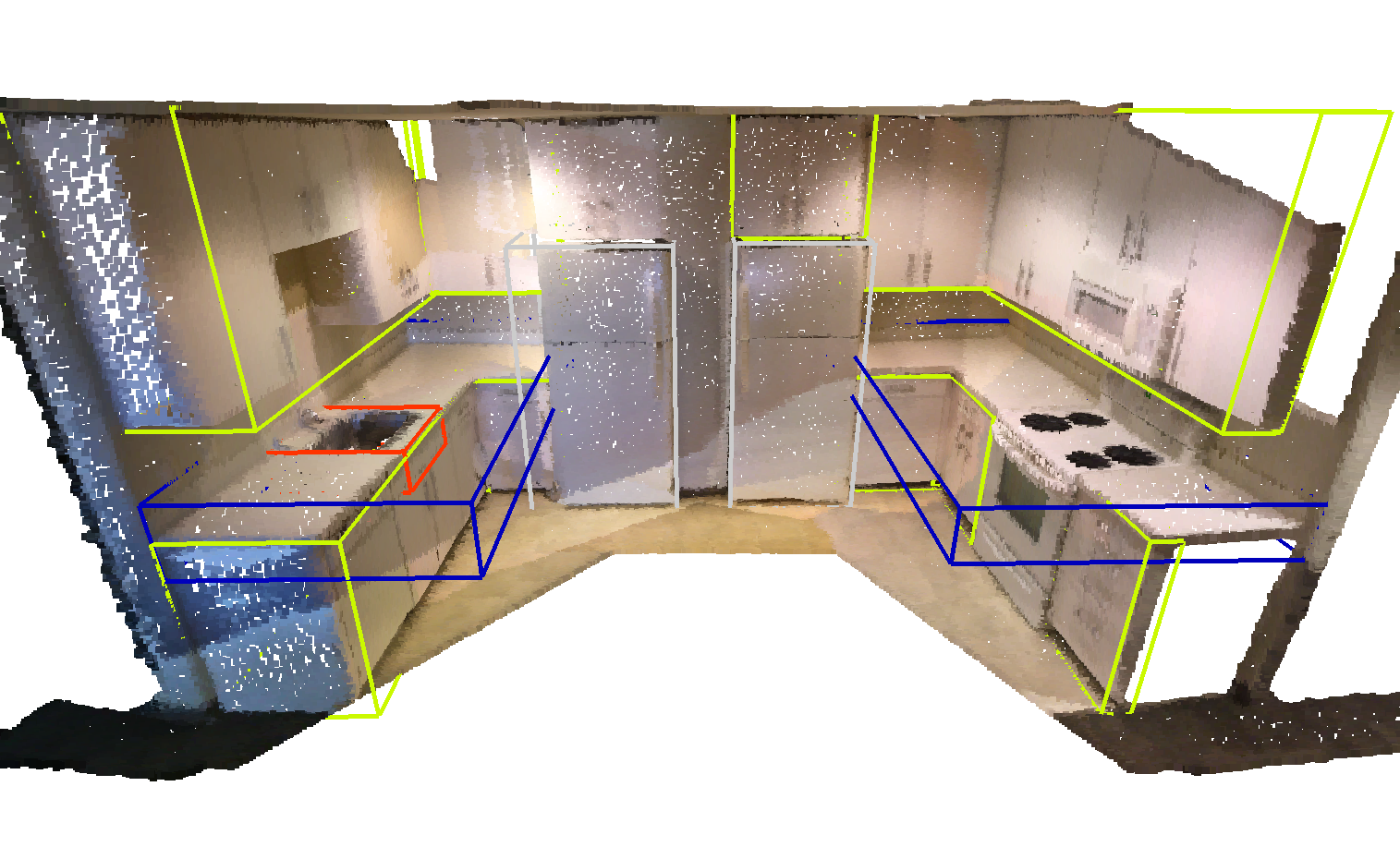} &
        \includegraphics[width=0.475\linewidth]{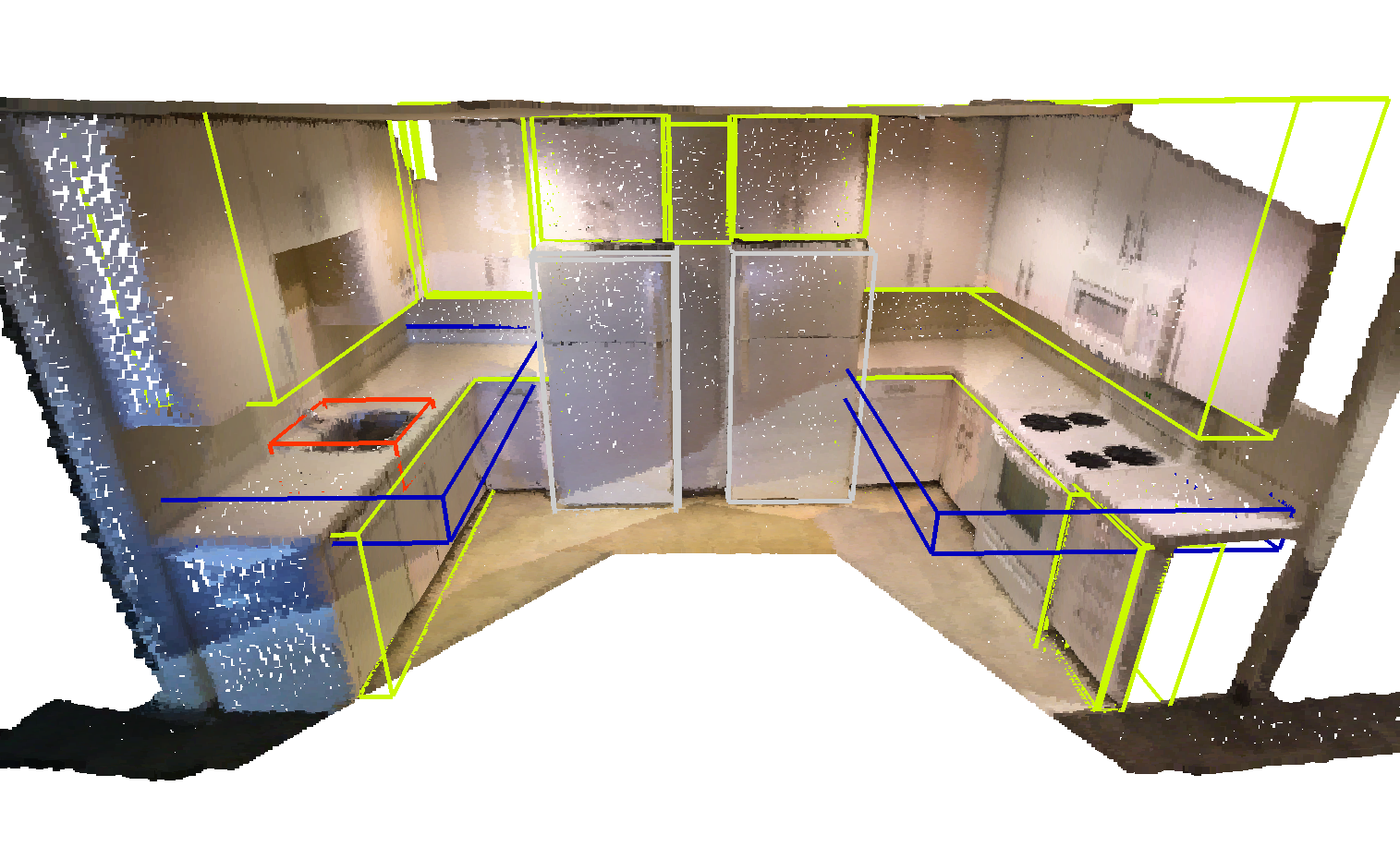} \\
        \parbox[t]{2mm}{\rotatebox[origin=l]{90}{\hspace*{3mm}SUN RGB-D}} & 
        \includegraphics[width=0.5\linewidth]{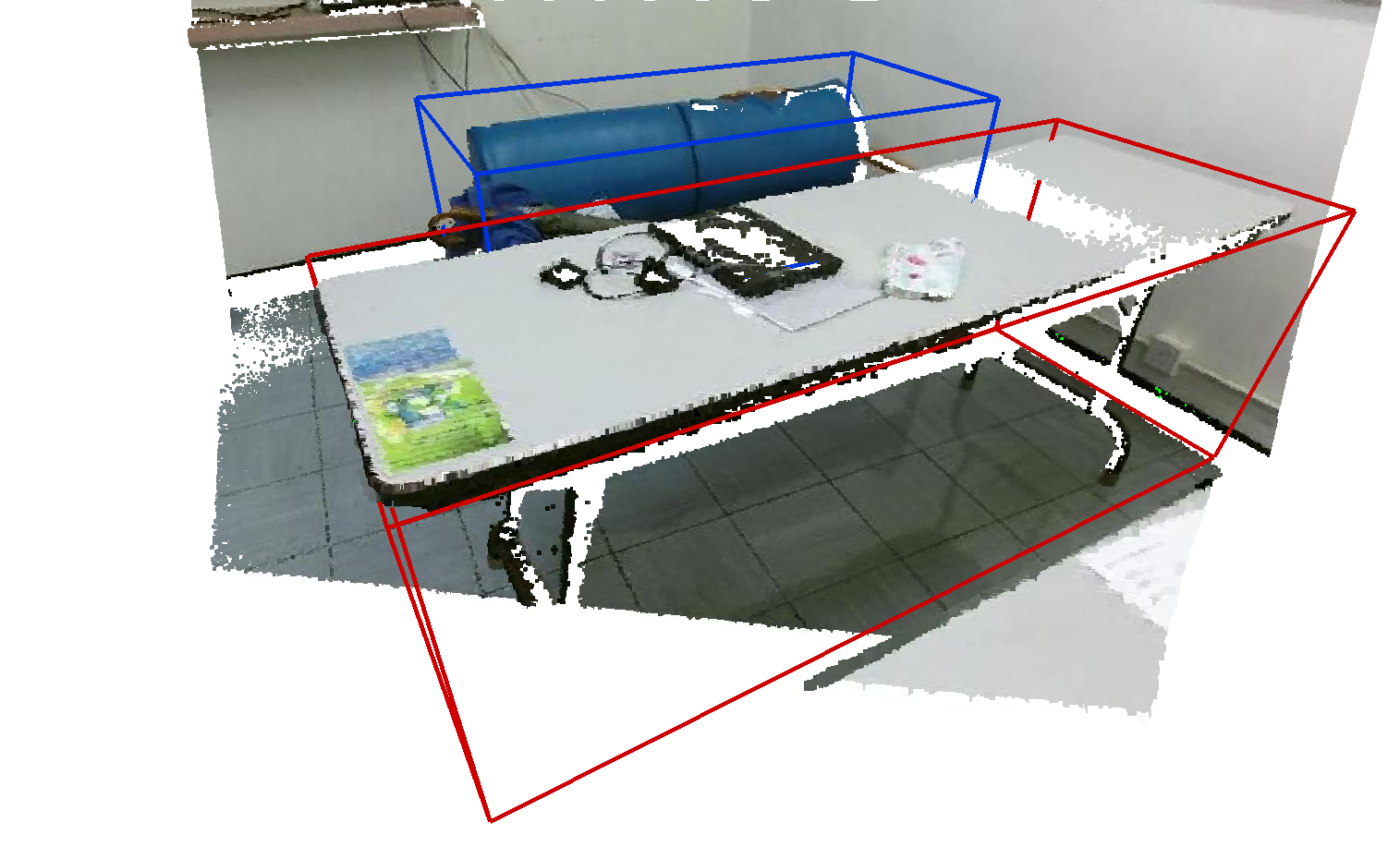} &
        \includegraphics[width=0.5\linewidth]{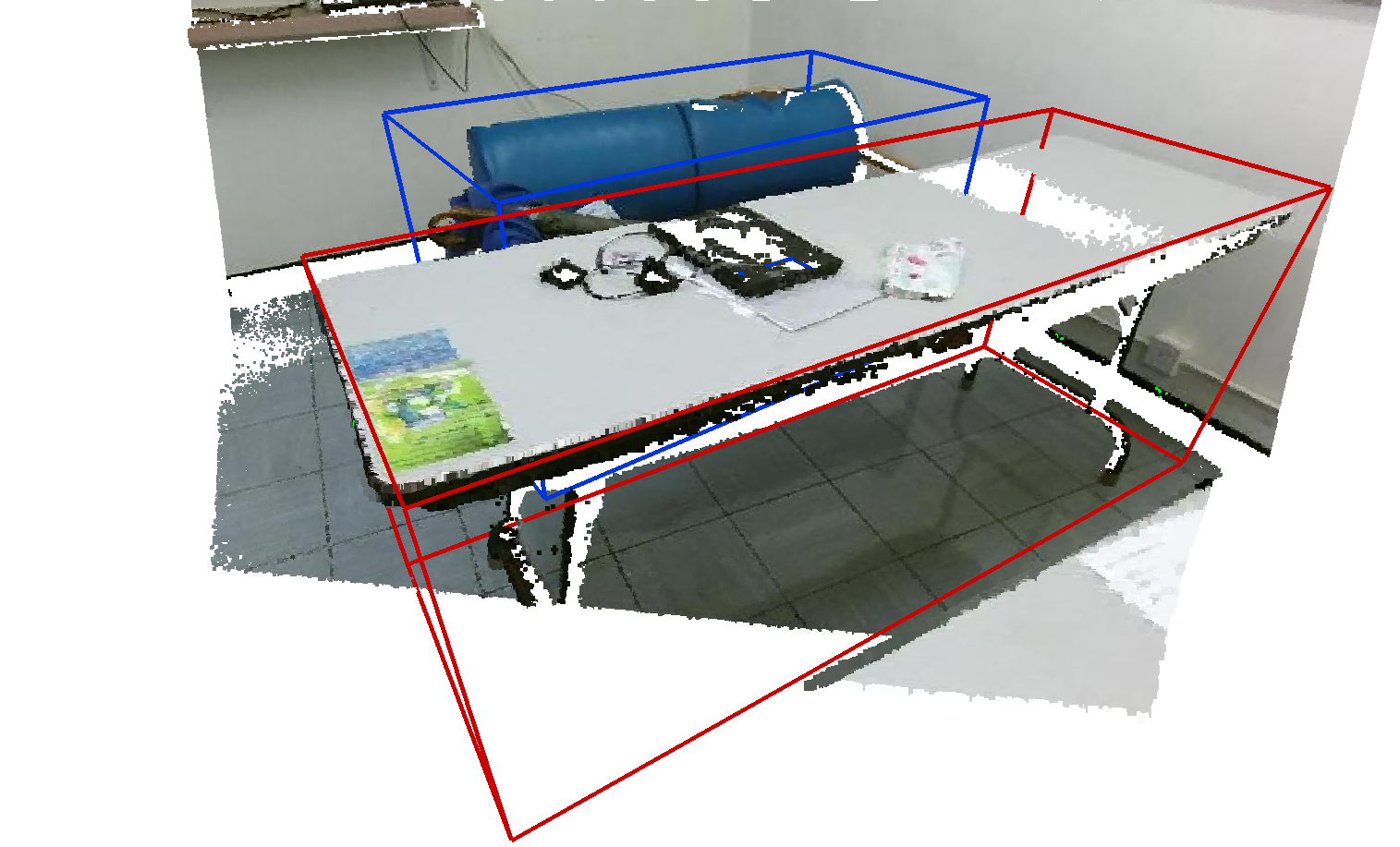} \\
        \parbox[t]{2mm}{\rotatebox[origin=l]{90}{\hspace*{6mm}S3DIS}} &
        \includegraphics[width=0.475\linewidth]{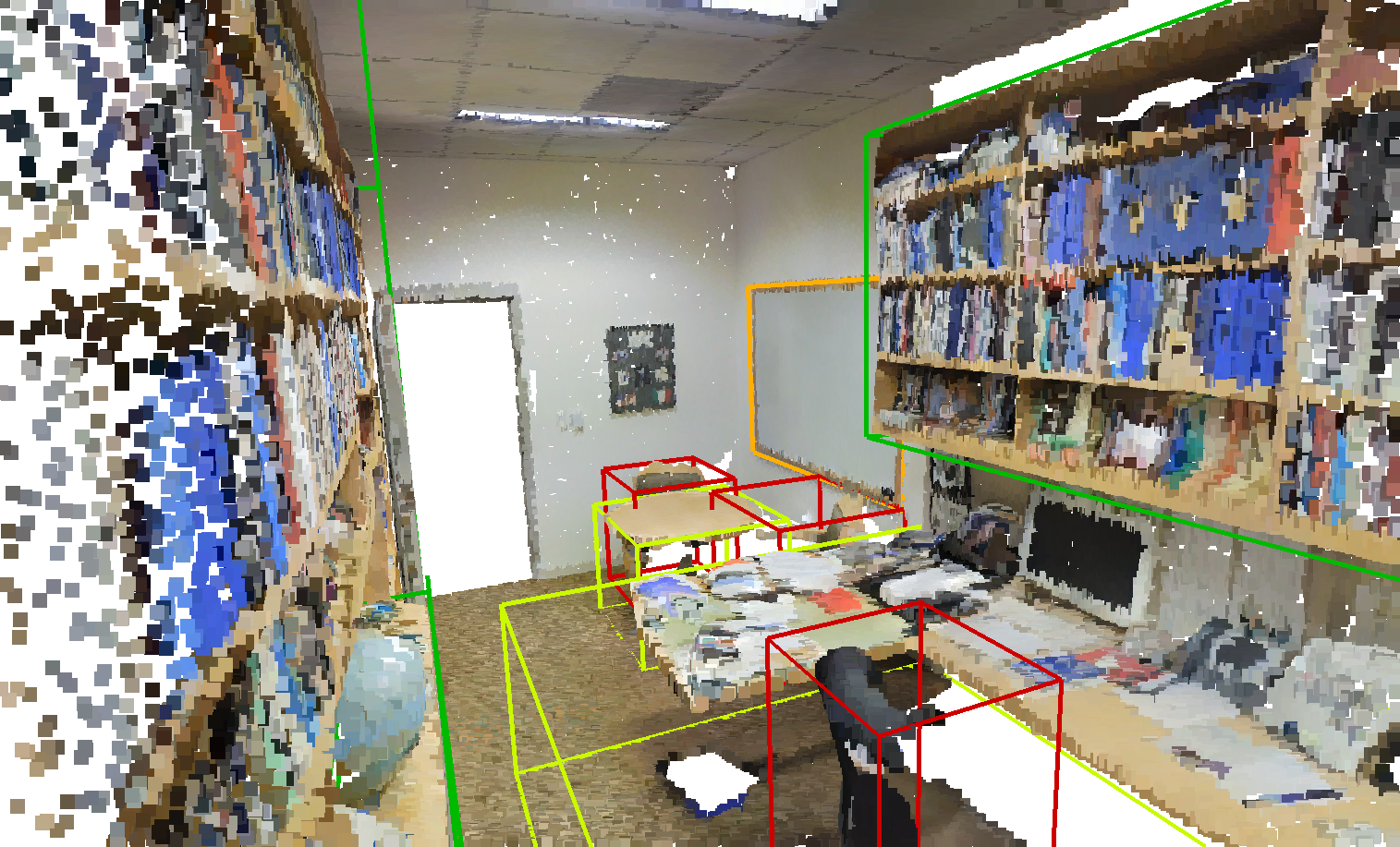} &
        \includegraphics[width=0.475\linewidth]{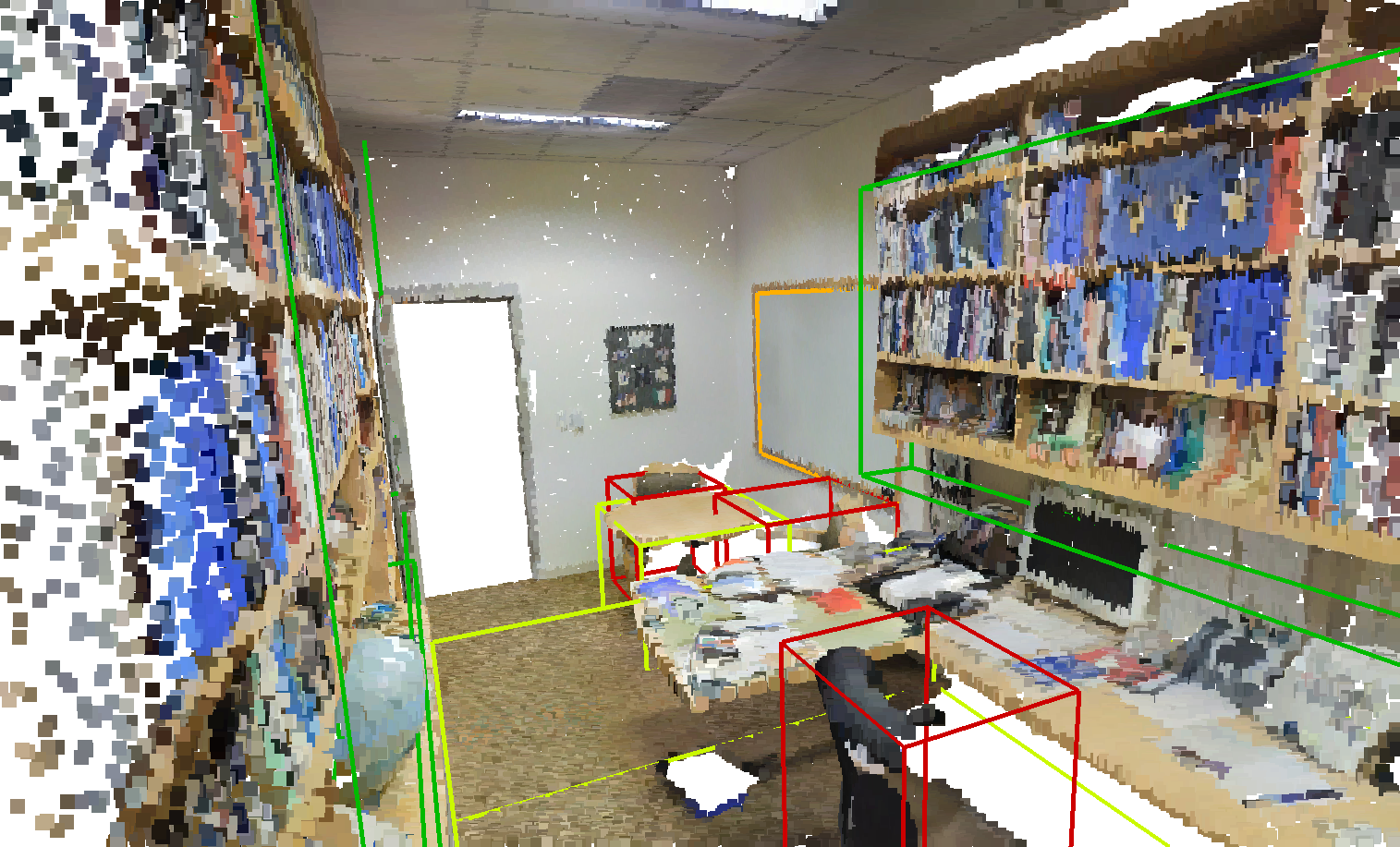} \\
    \end{tabular}
    \captionsetup{labelformat=empty}
    \captionof{figure}{\textbf{Figure 2}: Ground truth objects and objects detected by our \ours{} in the ScanNet, SUN RGB-D, and S3DIS point clouds.}
    \label{fig:qualitative}
\end{table}

\section{Conclusion}

In this paper, we introduced \ours{}, a novel 3D object detection method. Moreover, we developed an early fusion strategy to incorporate visual features into a 3D processing pipeline and proposed a modification of \ours{} called \oursff, that leverages both point cloud and RGB inputs. We evaluated the proposed methods on the standard benchmarks: ScanNet v2, SUN RGB-D, and S3DIS. Through experiments, we demonstrated that \ours{} outperforms existing methods in both accuracy and speed, thereby setting a new state-of-the-art in point cloud-based 3D object detection, while \oursff{} achieves the best results among methods using RGB and point clouds. 

\vfill\pagebreak

\clearpage

\bibliographystyle{IEEEbib}
\bibliography{refs}

\end{document}